\documentclass[10pt,twocolumn,letterpaper]{article}

\usepackage{cvpr}
\usepackage{times}
\usepackage{epsfig}
\usepackage{graphicx}
\usepackage{amsmath}
\usepackage{amssymb}

\usepackage[pagebackref=true,breaklinks=true,letterpaper=true,colorlinks,bookmarks=false]{hyperref}

\cvprfinalcopy 

\def\cvprPaperID{1587} 

\def\Im{\mathcal{I}}
\def\Loss{\mathcal{L}}

\def\s{{\bf s}}
\def\Re{\mathbb{R}}
\def\prob{\mathbb{P}}
\def\hIm{\hat{\mathcal{I}}}

\def\eE{\mathbb{E}}
\begin{document}

\title{Triple consistency loss for pairing distributions in GAN-based face synthesis}

\author{Enrique Sanchez and Michel Valstar \\
Computer Vision Laboratory, The University of Nottingham\\
{\tt\small \{Enrique.Sanchezlozano, Michel.Valstar\}@nottingham.ac.uk}
}

\maketitle


\begin{abstract}
\noindent Generative Adversarial Networks have shown impressive results for the task of object translation,  including face-to-face translation. A key component behind the success of recent approaches is the  self-consistency loss, which encourages a network to recover the original input image when the output generated for a desired attribute is itself passed through the same network, but with the target attribute inverted. While the self-consistency loss yields photo-realistic results, it can be shown that the input and target domains, supposed to be close, differ substantially. This is empirically found by observing that a network recovers the input image even if attributes other than the inversion of the original goal are set as target. This stops one combining networks for different tasks, or using a network to do progressive forward passes. In this paper, we show empirical evidence of this effect, and propose a new loss to bridge the gap between the distributions of the input and target domains. This ``triple consistency loss'', aims to minimise the distance between the outputs generated by the network for different routes to the target, independent of any intermediate steps. To show this is effective, we incorporate the triple consistency loss into the training of a new landmark-guided face to face synthesis, where, contrary to previous works, the generated images can simultaneously undergo a large transformation in both expression and pose. To the best of our knowledge, we are the first to tackle the problem of mismatching distributions in  self-domain synthesis, and to propose ``in-the-wild'' landmark-guided synthesis. Code will be available at \url{https://github.com/ESanchezLozano/GANnotation}\footnote{\url{https://youtu.be/-8r7zexg4yg}}.
\end{abstract}

\begin{figure}[t!]
\centering
  \includegraphics[width=0.95\columnwidth]{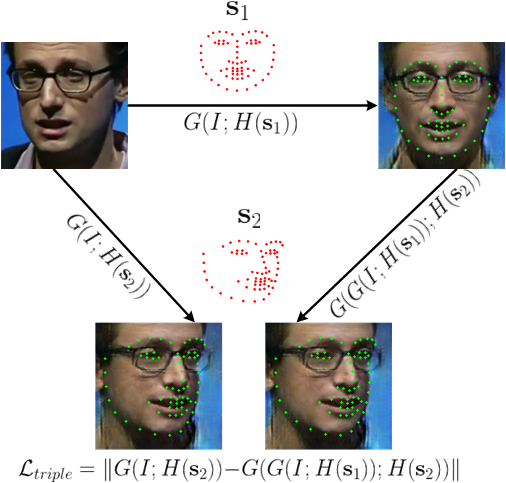}
  \caption{A triple consistency loss favours networks to produce similar results irrespective of their path to the target; either directly to $\s_2$ or via $\s_1$. $G$ is the generator function, $I$ is the input image, and $H(\s_t)$ are the heatmaps defined by the target shape $\s_t$. }
  \label{fig:triplelossabstract}
\end{figure}

\section{Introduction}
\begin{figure*}[t!]
    \centering\includegraphics[width=.95\linewidth]{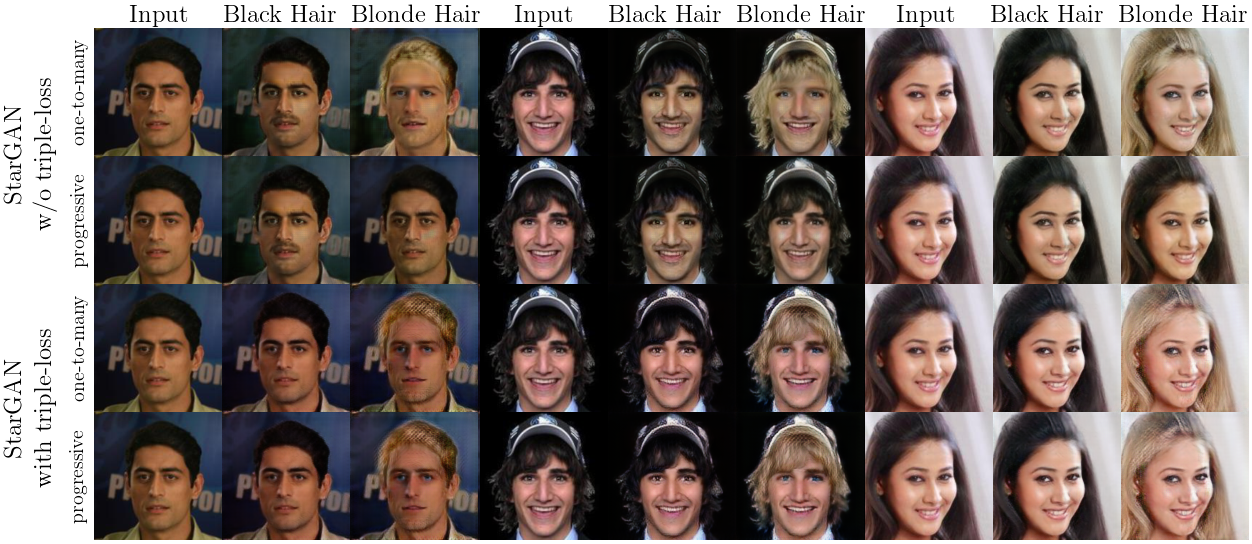}\par
  \caption{This paper illustrates the drawback of the self-consistency loss for identity preserving. The top two rows show the results of the original StarGAN~\cite{Choi2018} when the input image is translated into the attributes ``Black Hair'' or ``Blonde Hair''. The second row illustrates the result of generating the ``Blonde Hair'' attribute using as input the image generated from the ``Black Hair'' column. We can see that the output of the ``Blonde Hair'' recovers the input image with just subtle changes in the contrast, i.e. it fails to generate the target attribute. In the bottom rows, we illustrate the same process where the StarGAN is trained using the triple consistency loss presented in this paper. It can be seen that now the new attributes are correctly placed even after a first pass to the network.}
  \label{problem}
\end{figure*}


\noindent Recent advances in Generative Adversarial Networks (GANs~\cite{Goodfellow2014}) have found a broad range of applications in the domain of face synthesis or face-to-face translation~\cite{Huang2017,Choi2018,Di2018,Kossaifi2018,Pumarola2018}. Most face-to-face synthesis scenarios translate a target set of attributes~\cite{Choi2018}, landmarks~\cite{Wang2018}, or expressions~\cite{Pumarola2018}, onto the face present in the input image, where an additional goal is to preserve the  identity of the input face. One would expect the generated images to follow a similar probability distribution as that of the input images. However, while the generated images can be said to be photo-realistic, the distributions generated by the currents state of the art differ in an important way from the corresponding input domain. Upon closer inspection, we reveal an interesting phenomena: when the generated images are re-introduced to the network with a new set of target attributes, the network yields poor results, and occasionally fails to produce even photo-realistic images. We will refer to this way of generating images one after another as ``progressive image translation'', or simply \textit{progressive}.\\ 
\indent This major problem has remained unnoticed mainly due to the fact that existing approaches deploy a one-to-many image translation, where the target domain is always overlaid onto the input image. However, the problem leaves existing approaches with no hope solving the goal of achieving step-wise attribute translation, i.e. progressive image translation. Consider the following example goal: \textit{Can we use a network to convert the hair of a given person in an image from blonde to brown, and then use another network to modify their corresponding facial expression?} With the flaws of the current methods based on the self-consistency loss, the answer is no. \\
\indent In this paper, we argue that the reason behind this mismatch rises from the recently introduced self-consistency loss, where the network is expected to recover the input image from the generated one, if the inverted attribute is set as the target. This loss is used to enforce the network to preserve identity. However, we observe that when this condition is met, the input image is well recovered, no matter what target attribute is given back to the network, i.e. it appears that the network leaves a footprint in the generated image, not perceptible to the human eye, but that is evident when the generated image is re-introduced to the network. This is illustrated in Fig.~\ref{problem} (second row), where we show the effect of using the pre-trained StarGAN~\cite{Choi2018} to progressively generate the ``Blonde Hair'' attribute after first having generated the ``Black Hair" attribute. \\
\indent Having first discovered this problem, this paper presents a first approach to tackle this problem by introducing a new consistency loss, which we coin \textit{triple consistency loss}. This triple consistency loss (Fig.~\ref{fig:triplelossabstract}) aims to bridge the gap between the input and target distributions by imposing that any generated image has to be the same no matter if it is targeted by the network in one step or two. After retraining the StarGAN network with our triple consistency loss, we can see that the ``Blonde Hair'' attribute is correctly placed even when using as input the output of the network after generating the ``Black Hair'' attribute (Fig.~\ref{problem}, bottom row). In addition, we present our novel approach to unconstrained landmark guided face-to-face synthesis, which we name \textit{GANnotation}, and use this to illustrate the efficacy of the triple consistency loss. \\
\indent GANnotation translates a given face to a set of target landmarks, given to the network in the form of heatmaps. To the best of our knowledge, our GANnotation is the first network that allows synthesising faces in a wide range of poses and expressions, and can be used to construct person-specific datasets with very little supervision. An example is depicted in Fig.~\ref{fig:triplelossabstract}, where the input image $I$ is translated into the target point configurations $\s_1$ (and $\s_2$). We show that the target points become the ground-truth at the generated images, thus being practical for face alignment applications. We will release our code and models for the community to construct their own datasets as well as to encourage further research in this topic.

In short, the contributions of this paper can be summarised as follows:
\vspace{-0.1cm}
\begin{itemize}
\item We propose a \textbf{triple consistency loss} to bridge the gap between the distributions of the input and generated images. This enables the training of networks that not only reproduce photo-realistic images, but are also suitable for its use in combination with other networks. To the best of our knowledge, we are the first to introduce a triple consistency loss, which better represents the target distribution, allowing  progressive image translation.
\item We propose \textbf{GANnotation}, the first network that applies a face-to-face synthesis with simultaneous change in pose and expression. GANnotation is a network that can synthesise faces for a set of unconstrained target landmark annotations, whereby the given landmarks correspond to the ground-truth points in the generated images. 
\end{itemize}


\section{Related Work}
\noindent Generative Adversarial Networks (GANs)~\cite{Goodfellow2014} have proven to be a powerful tool in many Computer Vision disciplines, such as image generation~\cite{Radford2015}, style transfer~\cite{Johnson2016}, or super-resolution~\cite{Ledig2017}. 
In the context of image to image translation~\cite{Isola2017,Zhu2017}, GANs are composed of a generator that aims to reproduce the target domain, and a discriminator that tells whether the output of the generator is close to the target distribution or not. Both are learnt simultaneously using the minimax strategy. Since the introduction of GANs, many improvements on  adversarial learning have been proposed, including the Least-Squares GAN~\cite{Mao2017}, the Wasserstein GAN~\cite{Gulrajani2017,Arjovsky2017}, the Geometric GAN~\cite{Lim2017}, or  Spectral Normalisation~\cite{Wang2018non,Zhang2018,Miyato2018}, however there is no consensus as to which exhibits a systematic improvement. 

Reports on works suggesting improvements to the state of the art in GANs are often applied to the face domain. We review those that we consider the closest to our proposed approach, which aims to generate faces conditioned to attributes, landmarks, or expressions. There are works that proposed to do face frontalisation (\textbf{TP-GAN}~\cite{Huang2017}, \textbf{FF-GAN}~\cite{Yin2017}), profile face synthesis (\textbf{DA-GAN}~\cite{Zhao2017}), or multi-view image generation (\textbf{CR-GAN}~\cite{Tian2018}). While they employ different strategies, they have a common goal of preserving identity. However, these methods do not allow the synthesis of customised expressions or poses, so while CR-GAN can generate $9$ different views, these can not include additional synthesised expressions. We will show how our GANnotation \emph{can} perform both tasks with a landmark-guided synthesis.\\  

There exist other works that have proposed landmark-guided synthesis from a random seed, but without the aim to preserve identity. For instance, \textbf{GP-GAN}~\cite{Di2018} attempts to generate landmark-guided samples that are only conditioned on gender, making this method limited to variations in expression only. Furthermore, no variations in pose are shown. A similar approach is \textbf{GAGAN}~\cite{Kossaifi2018}, an appearance-preserving face generator that generates a random image  from a latent space and a target shape. Both the GP-GAN and GAGAN generate random faces, and thus cannot be used for person-specific data augmentation. Similarly, the \textbf{CMN-Net}~\cite{Wang2018} is a landmark guided smile generator, where a landmark to image synthesis network is used to generate frontal images under different expressions, supported by a recurrent neural network to preserve spatial consistency in the landmark generation. However, this method is limited to frontal faces, and the input image is expected to be neutral. Also, the \textbf{GC-GAN}~\cite{Qiao2018} is a geometry-aware method that is used to synthesise images from landmarks, where these are meant to display expressions. However, this method does not account for changes in pose. In contrast, the \textbf{CAPG-GAN}~\cite{Hu2018} applies pose-specific face rotation, where the input and target pose are encoded in sets of five heatmaps each, so that the network can perform attention. The five points are meant to capture head pose, which as an unwanted side-effect removes the network's ability to perform expression synthesis. The problem of expression synthesis was recently approached by \textbf{GANimation}~\cite{Pumarola2018}, where the generated images undergo a translation in the displayed expression. However, this method does not allow changes in pose, and it is limited to near-frontal faces. Finally, aside from expression and pose synthesis works, it is worth mentioning the \textbf{StarGAN}~\cite{Choi2018}, a multiple domain image-to-image translation approach, that allows changing facial attributes and expressions in a given image. This method is also limited to appearance changes, and thus does not tackle changes in pose or expression. 

Most of the  methods mentioned above apply a self-consistency loss to preserve identity. As shown in Fig.~\ref{problem}, this loss limits existing methods to  one-to-one mappings, and render generated images that are unable to be used as the basis for generating further images. These methods might leave a neutral face to a given expression~\cite{Pumarola2018}, or a non-frontal face to a frontal one~\cite{Huang2017}. In either case, the network is not required to perform more than one forward pass from a given image. Thus, a self-consistency loss is applied to either preserve appearance or identity. While this yields impressive results, it causes a mismatch between the input and target distributions, when a desired property would be to actually make them match. 
As we shall see, our proposed GANnotation does have this property, and thanks to that \textit{GANnotation is the first method that can generate faces with a target pose and expression simultaneously}.




\section{Proposed approach}
\begin{figure}[t!]
\centering
  \includegraphics[width=0.97\columnwidth]{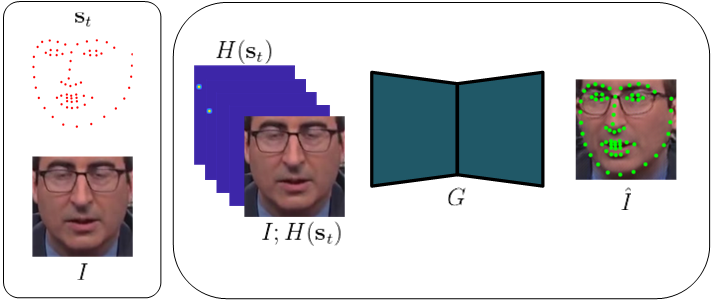}
  \caption{Proposed approach: We are given an input image $I$, and a set of target points $\s_t$. The points are encoded as a set of heatmaps $H(\s_t)$ and concatenated with the input image $I;H(\s_t)$. The concatenated volume is sent to the generator $G$ to produce the image $\hat{I}$, we overlay the target points on the generated image to illustrate the main task of the network.}
  \label{fig:network}
\end{figure}

\noindent Our goal is to generate (synthesise) a set of person-specific images driven by a set of landmarks, so that these become the ground-truth landmarks in the generated image. Contrary to previous works, we want our network to allow for simultaneous changes in both pose and expression. In addition, we want the generated images to be not only photo-realistic, but also to be distribution-wise close to the input images. To the best of our knowledge, this is the first work that directly permits changes in pose and expression simultaneously, and that reduces the gap between the input and target distributions. An overall description of our proposed approach is depicted in Fig.~\ref{fig:network}.

\subsection{Notation}
\label{sec:notation}
\noindent Let $I \in \mathcal{I}$ be a $w \times h$ pixels face image, for which a set of $n$ indexed points $\s_i \in \Re^{2n}$ is available. $\mathcal{I}$ represents the space of original images of size  ${w \times h}$. The \textit{generator} is a function $G: \mathcal{I} \times \mathcal{H} \rightarrow \hat{\mathcal{I}}$, with $\hat{\mathcal{I}}$ the space of generated images, that receives as input an image $I$ and a set of heatmaps $H(\s_t) \in \mathcal{H}$ encoding a target shape $\s_t$, and outputs the warped image. In particular, the estimated image $\hat{I}$ is defined as:
\begin{equation}
\hat{I} = G\left( I ; H(\s_t) \right)
\end{equation}
where $;$ indicates that $I$ and $H$ are concatenated. The notation $H(\s)$ is used to represent the dependence of the heatmaps w.r.t. the shape $\s$. In particular, $H(\s) \in \Re^{n \times w \times h}$ is defined as a set of $n$ heatmaps (one for each facial point), each itself being a $w \times h$ map, in which a unit Gaussian is centred at its corresponding landmark. 
In general, we will assume that images $I$ are drawn from a \textit{real} distribution $\prob_\Im$, and that generated images $\hat{I}$ are said to be drawn from $\prob_{\hIm}$. In some scenarios, like the one presented in this paper, we want $\prob_{\hIm}$ to match $\prob_\Im$ as closely as possible. 
The discriminator will be defined as a function $D$ that receives as input an estimated image $\hat{I}$, or a real image $I$, and aims to label them as real or fake. 

\subsection{Architecture}
\noindent The generator is adapted from the architecture successfully proposed for the task of neural transfer~\cite{Johnson2016}, and later adapted to the image-to-image translation task~\cite{Isola2017,Zhu2017}. This architecture has also proven successful for the task of face synthesis~\cite{Pumarola2018,Ma2017,Hu2018}, and basically consists of two spatial downsampling convolutions, followed by a set of residual blocks~\cite{He2016}, and two spatial upsampling blocks with $1/2$ strided convolutions. The generator is modified to account for the $ 3 + n $ input channels, defined by the RGB input image and the heatmaps corresponding to the target landmark locations. As in~\cite{Pumarola2018}, we adopt a mask-based approach, by splitting the last layer of the generator into a colour image $C$ and a mask $M$. The output of the generator is thus defined as:
\begin{equation}
\hat{I} = (1 - M) \circ C + M \circ I,
\end{equation}
where $\circ$ represents an element-wise product. Without loss of generality, we refer to $\hat{I}$ as the output of the generator. Further details of the network can be found in the main project site. 

The discriminator is adopted from the PatchGAN~\cite{Isola2017,Zhu2017} architecture, and consists of several convolution-based downsampling blocks, each increasing the number of channels to $512$, and followed by a LeakyReLU~\cite{Maas2013}. For an input resolution of $128\times 128$ this network yields an output volume of $4 \times 4 \times 512$, which is forwarded to a FCN to give a final score. 



\subsection{Training}
\noindent The loss function we aim to optimise consists of seven terms. Below we give a mathematical formulation for each, and introduce the triple-consistency loss, which is our main contribution.

\subsubsection{Adversarial loss} 
\noindent We adopt the hinge adversarial loss proposed in~\cite{Lim2017}, which is shown to require less updates in the discriminator per update in the generator, also allowing a faster learning~\cite{Zhang2018,Miyato2018}. The loss for the discriminator is defined as:
\begin{eqnarray}
\mathcal{L}_{adv}(D) = &-&\eE_{\hat{I}\sim \prob_{\hat{I}}} [ \min(0,-1+D(\hat{I}))] \nonumber \\
&-& \eE_{I\sim \prob_I} [\min(0,-D(I)-1) ],
\end{eqnarray}
whereas the loss for the generator is defined as:
\begin{equation}
\mathcal{L}_{adv}(G) = -\eE_{I\sim \prob_I} [  D(\hat{I}) ].
\end{equation}

\subsubsection{Pixel Loss} 
\noindent In order to make the network learn the target representation, we use a pixel reconstruction loss. In particular, for a given input image $I$ and target points $\s_t$, corresponding to the ground-truth points of a ``target" image $I_t$, the pixel loss is defined as\footnote{For the sake of clarity, we will onward omit the expectation term}:
\begin{equation}
\Loss_{pix} = \| G(I ; H(\s_t)) - I_t \|_2^2.
\end{equation}
This loss is used along with a \textit{total variation regularisation} loss, $\Loss_{tv}$~\cite{Aly2005,Johnson2016}, which encourages the output of the network to yield smoothness in the generated images. 

\subsubsection{Consistency Loss} 
\noindent In the context of face to face synthesis, the generator is expected to be able to \textit{invert} the transform applied to the input image. In practice, this is accomplished by feeding the generator with its output when a given image with target points is given. This loss is also referred to as the identity loss in~\cite{Pumarola2018,Choi2018}, and is defined as:
\begin{equation}
\Loss_{self} = \| G\left( G( I ; H(\s_t) ) ; H( \s_i ) \right) - I \|^2.
\label{selfloss}
\end{equation}
where $\s_i$ represents the ground-truth points for the image $I$. In practice, the consistency loss is obtained by first passing the input image with the target landmarks to the generator, and then by passing the corresponding output with the initial landmarks back to the generator. 
\subsubsection{Triple Consistency Loss}
The self-consistency loss shown above was presented in~\cite{Pumarola2018,Choi2018} to enforce the network to preserve identity. This  means that the network will recover the original image when the original expression is given as a target to the output of a first pass. In~\cite{Pumarola2018,Choi2018} this approach is specifically defined as in Eq.~\ref{selfloss}. However, we have noticed that this loss causes the network to recover the input image \textit{no matter what further target is considered}, when we expect this to only happen with the further target set as the inverse of the original target. 

Rather than capturing the input distribution, the network translates images into a domain that encodes the input image along with the output, i.e., $\prob_{\hIm} \nsim \prob_\Im$. We conjecture that this problem has so far remained undiscovered due to the fact that existing works set a neutral-to-expression synthesis goal rather than expression-to-expression, which means that the input and output spaces do not need to overlap. However, we want the network to not only produce photo-realistic images, but also to make them reusable, and therefore the input and output domains need to be similar. 

In order to solve this problem, and allow progressive image generation, we introduce the triple-consistency loss. In particular, when an image is sent to a target location, and its output is re-sent to another location, we expect the network to also do so in a single pass. Given the input image $I$, and target points $\s_t$, the output of the generator is $\hat{I} = G( I ; H(\s_t) )$. Now, we observe that sending $I$ and $\hat{I}$ to another target location $\s_n$ should result in similar outputs. That is to say, we want $G( \hat{I} ; H(\s_n) )$ to be similar to $G(I ; H(\s_n))$. The triple-consistency loss is thus defined as:
\begin{equation}
\Loss_{triple} = \| G( \hat{I} ; H(\s_n) ) - G( I ; H(\s_n) ) \|^2
\end{equation}
The overall idea of the triple consistency loss is depicted in Fig.~\ref{fig:triplelossabstract}. This loss will try to enforce $\prob_{\hIm} \sim \prob_\Im$. 

\subsubsection{Identity preserving loss}
In order to enforce the network to preserve the identity wherever the target points allow the generated image to do so, we also use the identity preserving network, coined Light CNN, presented in~\cite{Wu2018}. We use a similar approach to~\cite{Hu2018,Huang2017} and define the identity loss as the $l_1$ norm between the features extracted at the last two layers of the Light CNN w.r.t. both the generated and the real images. In particular, denoting $fc$ and $p$ as the fully connected layer and last pooling layer of the Light CNN network, respectively, and $\Phi^{l}_{CNN}$ the features extracted at the layer $l=\{fc,p\}$, the identity loss is defined as:
\begin{equation}
\Loss_{id} = \sum_{l=fc,p} \| \Phi^{l}_{CNN}(I) - \Phi^{l}_{CNN}(\hat{I}) \|
\label{identity}
\end{equation}

\subsubsection{Perceptual loss}
In order to provide the network with the ability to generate subtle details, we follow the line of recent approaches in super resolution and style transfer~\cite{Ledig2017,Bulat2018}, and use the perceptual loss defined by~\cite{Johnson2016}. The perceptual loss enforces the features at the generated images to be similar to those of the real images when forwarded through a VGG-19~\cite{Simonyan15} network. The perceptual loss is split between the feature reconstruction loss and the style reconstruction loss. The feature reconstruction loss is computed as the $l_1$-norm of the difference between the features $\Phi^{l}_{VGG}$ computed at the layers $l = \{\mathtt{relu1\_2}, \mathtt{relu2\_2}, \mathtt{relu3\_3},\mathtt{relu4\_3}\}$ of the input and generated images. The style reconstruction loss is computed as the Frobenius norm of the difference between the Gram matrices, $\Gamma$, of the output and target images, computed from the features extracted at the $\mathtt{relu3\_3}$ layer:
\begin{eqnarray}
\Loss_{pp} = \sum_{l} \| \Phi^{l}_{VGG}(I) - \Phi^{l}_{VGG}(\hat{I}) \| \nonumber \\ + \| \Gamma(\Phi^{\mathtt{relu3\_3}}_{VGG}(I)) - \Gamma(\Phi^{\mathtt{relu3\_3}}_{VGG}(\hat{I})) \|_F.
\label{perceptual}
\end{eqnarray}
\subsubsection{Full loss}
The full loss for the generator is then defined as:
\begin{eqnarray}
    &\Loss(G) = \lambda_{adv} \Loss_{adv} + \lambda_{pix} \Loss_{pix} + \lambda_{self} \Loss_{self}  \nonumber \\
    &+ \lambda_{triple} \Loss_{triple} + \lambda_{id} \Loss_{id} + \lambda_{pp} \Loss_{pp} + \lambda_{tv} \Loss_{tv},
\end{eqnarray}
where, in our set-up, $\lambda_{adv} = 1$, $\lambda_{pix} = 10$, $\lambda_{self} = 100$, $\lambda_{triple} = 100$, $\lambda_{id} = 1$, $\lambda_{pp} = 10$, and $\lambda_{tv} = 10^{-4}$.

\section{Training Datasets}
\label{data}
\noindent Training the network requires the use of paired data, i.e. pairs of images from the same subject for which the points are known. However, we approach the training with triplets rather than pairs of images, in order to also be able to compare the output of the network after one and two passes with the ground-truth images. To this end, we use the training partition of the 300VW~\cite{Shen15}, which is composed of annotated videos of $50$ people. For each video, we choose a set of $3000$ triplets, where each triplet is composed of random samples from the video. In addition, we use the public partition of the BP4D dataset~\cite{Zhang2014bp4d,Valstar2015}, which is composed of videos of $40$ subjects performing $8$ different tasks. For each of the BP4D videos, we select $500$ triplets.

We found that using only $90$ subjects results in overfitting, which causes the network to lose its ability  to preserve identity. To overcome this problem, we augment our training set with unpaired data. In particular, we use a subset of $\sim$8000 images collected from datasets that are annotated in a similar fashion to that of the 300VW. We use Helen~\cite{Le12}, LFPW~\cite{Belhumeur13}, AFW~\cite{Zhu12}, IBUG~\cite{Sagonas13}, and a subset of MultiPIE~\cite{Gross10}. To ensure label consistency across datasets we used the facial landmark annotations provided by the 300W challenge~\cite{Sagonas13}. To generate triplets on this data, we apply random affine transformations to the images and points, as well as a random image mirroring. This makes every image to be ``paired'' with random affine perturbations on the landmarks. While the network will learn to translate non-rigid deformations from the 300VW subset, it will learn to preserve identity and be robust to rigid perturbations, including mirroring, from the subset of unpaired data.

\section{Experiments}
\noindent All the experiments are implemented in PyTorch~\cite{Paszke2017}, using the Adam optimiser~\cite{Kingma2015}, with $\beta_1 = 0.5$ and $\beta_2 = 0.9999$ . The input images are cropped according to a bounding box defined by the ground-truth landmarks with an added margin of 10 pixels each side, and then re-scaled to be 128x128. 

The model is trained for 30 epochs, each consisting of $10,000$ iterations, which takes approximately $24$ hours to be completed with two NVIDIA Titan X GPU cards. The batch size is $16$, and the learning rate is set to $10^{-4}$, and it is linearly decreased over 20 epochs to $10^{-6}$. The size of the heatmaps is $6$ pixels, corresponding to a unit 2D Gaussian. For each iteration a random batch is taken from either the paired or unpaired data, as described in Section~\ref{data}.




\begin{figure*}[t!]
\begin{center}
\begin{tabular}{|l|*{6}{c}|*{6}{c}|} \hline 
   & \multicolumn{6}{c|}{StarGAN without triple consistency loss}  & \multicolumn{6}{c|}{StarGAN with triple consistency loss} \\ \hline
   & \hspace{8pt} \scriptsize{Input} & \scriptsize{Black Hair} & \scriptsize{Blonde Hair} & \scriptsize{Brown Hair} & \scriptsize{Gender} & \scriptsize{Age} & \hspace{8pt} \scriptsize{Input} & \scriptsize{Black Hair} & \scriptsize{Blonde Hair} & \scriptsize{Brown Hair} & \scriptsize{Gender} & \scriptsize{Age} \\
  \rotatebox{90}{\scriptsize{Progressive | One-to-one}} & \multicolumn{6}{c|}{\includegraphics[width=0.98\columnwidth]{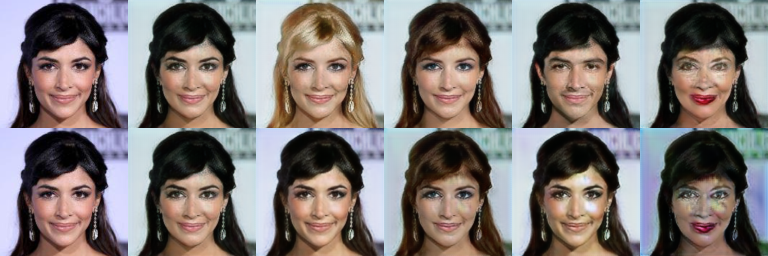}} &  \multicolumn{6}{c|}{\includegraphics[width=0.98\columnwidth]{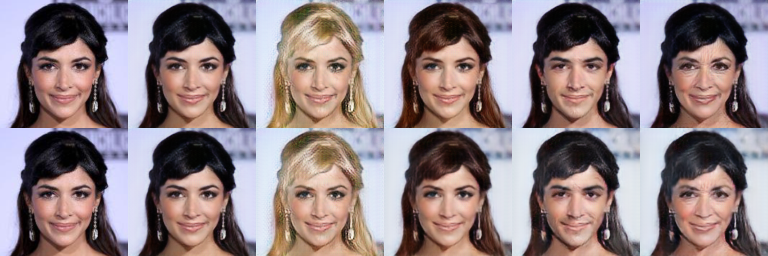}} \\ 
  \rotatebox{90}{\scriptsize{Progressive | One-to-one}} & \multicolumn{6}{c|}{\includegraphics[width=0.98\columnwidth]{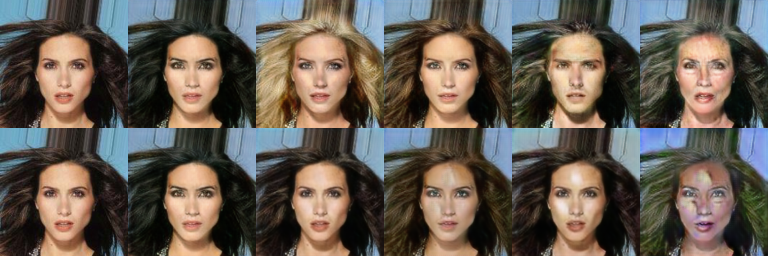}} & \multicolumn{6}{c|}{\includegraphics[width=0.98\columnwidth]{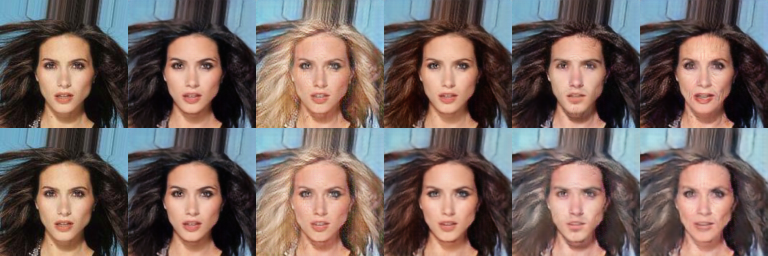}} \\ 
  \rotatebox{90}{\scriptsize{Progressive | One-to-one}} &
  \multicolumn{6}{c|}{\includegraphics[width=0.98\columnwidth]{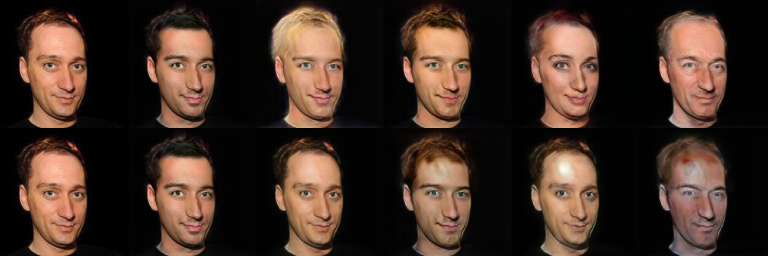}} & \multicolumn{6}{c|}{\includegraphics[width=0.98\columnwidth]{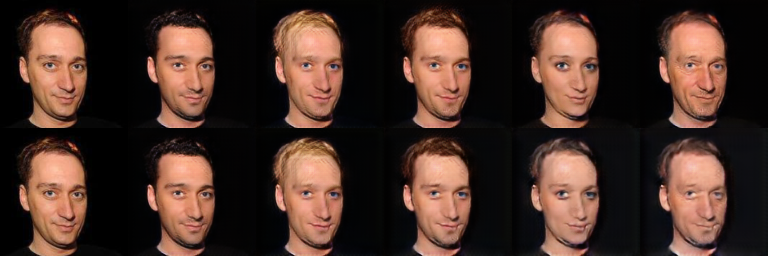}} \\ \hline
   \end{tabular}
  \end{center}
 
  \caption{Comparison between the original StarGAN~\cite{Choi2018} (left) w.r.t. a StarGAN trained with the triple consitency loss introduced in this paper. For the two examples illustrated, the first row corresponds to a one-to-one mapping where the input image is always used to generate the target attribute. The second row shows the results after using the output of the network as input for the next target attribute.}
  \label{fig:stargan}
\end{figure*}

\subsection{On the use of a triple consistency loss}
\noindent First, we want to validate the contribution of the triple consistency loss independently of our proposed approach. To do so, we re-use the StarGAN~\cite{Choi2018} implementation, as it is accompanied with author's trained model. We appended to the training the triple consistency loss and we compare the results of the retrained network with that provided by the corresponding authors. The original StarGAN model was trained on the Celeb-A dataset~\cite{Liu2015}, and it applies to a given face a set of attributes, namely ``Black Hair'', ``Blonde Hair'', ``Brown Hair'', ``Gender'', and ``Age''. The attributes of ``Gender'' and ``Age'' have to be understood as generating the opposite attribute to the one given in the input image. We show some results generated by the model in the first row of each example on Fig.~\ref{fig:stargan} and Fig.~\ref{problem}. In these examples, the same input image is used to generate all the target attributes. Then, using the same network, we apply a progressive image generation, whereby the output image after inserting the first attribute is forwarded to the network to create the second attribute, and so forth. In other words, the network takes as input the output of the network w.r.t. the previous attribute. The results of this progressive attribute translation are shown in the second row of Fig.~\ref{fig:stargan}.  We can see that the images degrade substantially as soon as the the network has to deal with a couple of generated images, generating burning-like artifacts. Then, \textit{we have re-trained the StarGAN network just including the triple consistency loss}, and repeated the same process as before. The corresponding results are shown in the bottom rows of Fig.~\ref{fig:stargan}. As it can be seen in the third row, the StarGAN trained with a triple consistency loss keeps a high-level image generation with the target attributes, while having a distribution that is closer to that of the input images. This is illustrated in the bottom row.
\begin{figure}[h!]
\centering
  \includegraphics[width=0.46\columnwidth]{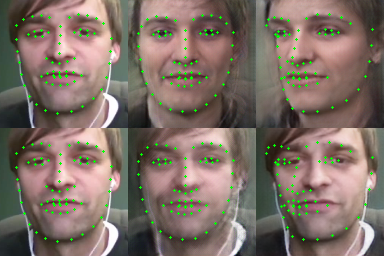} 
  \includegraphics[width=0.46\columnwidth]{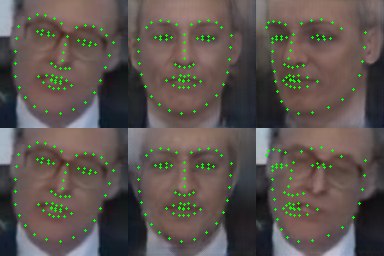}\\
  \includegraphics[width=0.46\columnwidth]{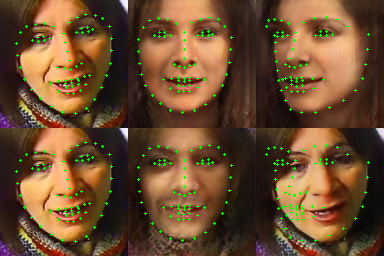}
  \includegraphics[width=0.46\columnwidth]{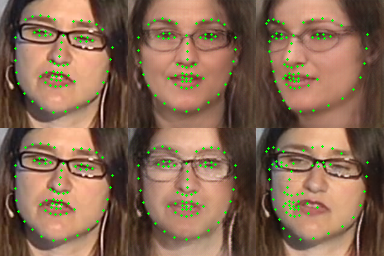} \\
  \includegraphics[width=0.46\columnwidth]{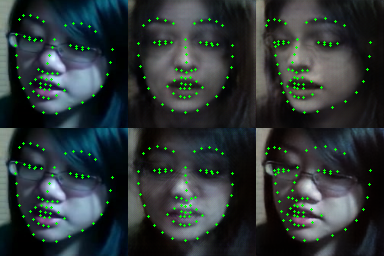}
  \includegraphics[width=0.46\columnwidth]{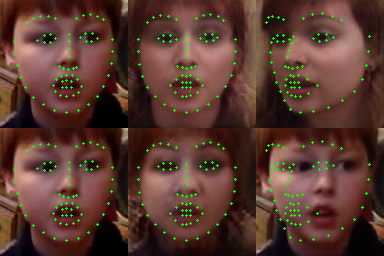} 
  \caption{Results attained by the model trained with a triple consistency loss (top rows) and a model trained without it (bottom rows). The left image corresponds to the input image, the middle image corresponds to the output of the network after the first pass, and the right image corresponds to the output of the network after a second pass. The points represent the target location.}
  \label{fig:onto}
\end{figure}\\

\indent Next, we show the contribution of the triple consistency loss within our GANnotation. We train two models under the same conditions, with, and without the triple consistency loss. At test time, we use a set of images from the test partition of 300VW~\cite{Shen15} for which there are available points. Each image is first frontalised using the given landmarks (see Section~\ref{gannot} for further details), and then sent to a pose-specific angle. The results are shown in Fig.~\ref{fig:onto}, where the top rows correspond to the images generated by a model trained with the triple consistency loss and the bottom rows represent the images generated by a model trained without the triple consistency loss. We show how after the first map, both images look alike, being similar to the input image. However, after the second pass, the generator trained without the triple consistency loss recovers the input images, with subtle changes in the contrast. This effect is not occurring with the images generated by the network trained with the triple consistency loss, where the images are correctly mapped. We also show how the network produces similar results after the first pass. We will release both models for further validation. As it can be seen, while both networks generate plausible images at a first pass, the former fails after subsequent forwards. 

\begin{figure*}[t!]
\centering
  \includegraphics[width=2.\columnwidth]{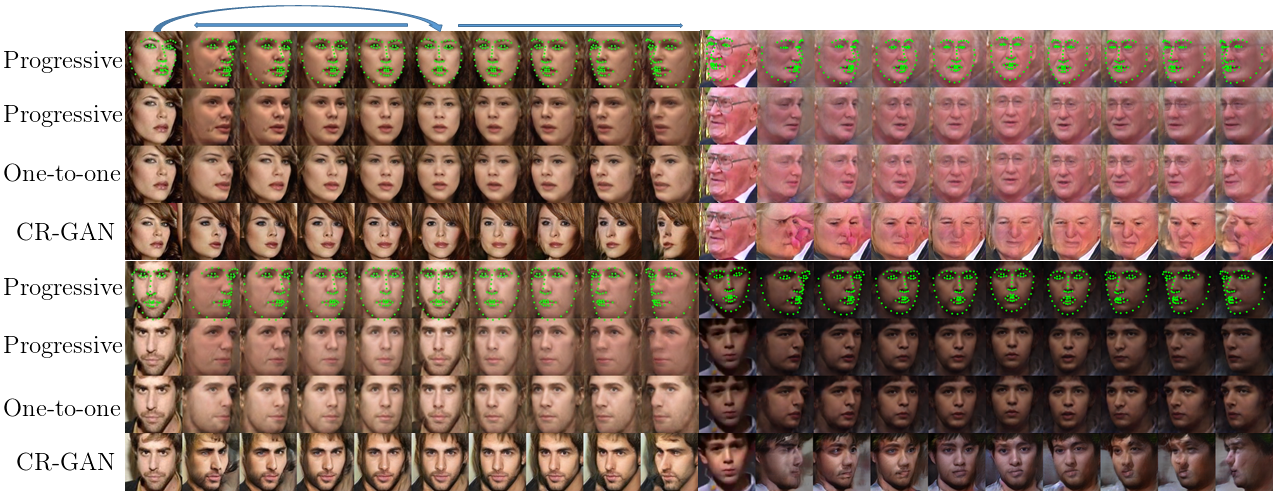}
  \caption{Landmark-guided multi-view synthesis and comparison with CR-GAN~\cite{Tian2018}. The first and second rows correspond to a progressive image generation (with and without the landmarks for a clear visualisation), whereby the input image (leftmost) is first frontalised (middle column), and then sent progressively to the corresponding views. The third row corresponds to a one-to-one mapping. The fourth row corresponds to the CR-GAN results. The two examples on the left images correspond to the images given along with the publicly available code of CR-GAN. We can see that our method yields realistic results despite the tight cropping, different to that used to train our GANnotation. On the right, two examples extracted from the 300VW test partition, cropped according to the landmarks. We can see that the CR-GAN fails to produce photo-realistic results.} 
  \label{fig:comparison_1}
\end{figure*}

\subsection{GANnotation}
\label{gannot}
\noindent We now evaluate the consistency of our GANnotation for the task of landmark-guided face synthesis. In order to compare our GANnotation w.r.t. most recent works, we apply a landmark-guided multi-view synthesis, and compare our results against the publicly available code of \textbf{CR-GAN}~\cite{Tian2018}. We compare our method in the test partition of the 300VW~\cite{Shen15}. To generate pose-specific landmarks, we use a shape model trained on the datasets described in Section~\ref{data}. The shape model includes a set of specific parameters that allow manipulating the in-plane rotation, as well as the view angle (pose). Using the shape model, we first remove both the in-plane rotation and the pose, resulting in the frontalised image given in the middle column. Then, the pose specific parameter is manipulated to generate the synthetic poses shown in the left and right columns w.r.t. the frontalised face. In addition, when generating the pose-specific landmarks, we randomly perturb the expresion related parameters, so as to generate different faces. The results are shown in Fig.~\ref{fig:comparison_1}. We show both the results of a \textit{progressive} image generation (first and second rows), as well as the one-to-one mapping (third row). Finally, we compare the results w.r.t. those given by the CR-GAN model. To show the performance of our GANnotation, we attach a video with a reenactment experiment\footnote{\url{https://youtu.be/-8r7zexg4yg}}, where the appearance of a given face is transferred to the points extracted from each frame of another video.   



\subsection{Remarks}
\noindent We have shown that our network yields photo-realistic results whilst maintaining certain consistency when applying multiple passes to the same network. In this Section, we want to remark an important aspect that needs consideration when using the triple consistency loss, as well as discuss to which extent the network will preserve the identity. \\ 
\indent \textbf{The effectiveness of the triple consistency loss}. This loss, when used with no self-consistency loss, can overcome the degradation problem completely. However, we have observed that when the self-consistency loss is removed from the training, the network is prone to failure at preserving identity. Therefore, while the triple consistency loss pulls the input image out of the target domain, the self-consistency loss is needed to better preserve identity. \\
\indent \textbf{Preserving identity vs. preserving the landmarks}. While our proposed approach can preserve identity in most cases, it is important to remark some cases where the network will likely fail: 1) When the target points force the network to do so. The network will generate plausible faces and will prioritise the target locations over the identity and even gender. An example is depicted in the most extreme views shown in Fig.~\ref{fig:comparison_1}, where the network enforces to locate the eyes where they are targeted, even when it means a less realistic face. Thus, if the target landmarks do not show identity consistency, the network will likely fail to preserve it. 2) When there is a big mismatch between the ground-truth points at the given image and the target landmarks. Given that the network is not provided with any attention mechanism, one of its tasks is to locate which information needs to be transferred to the target points. When the network fails to do so or the target points are displaced substantially from the input then identity can be poorly preserved.
\section{Conclusion}
\noindent In this paper, we have illustrated a drawback of face-to-face synthesis methods that aim to preserve identity by using a self-consistency loss. We have shown that despite images being realistic, they cannot be reused by the network for further tasks. Based on this evidence, we have introduced a triple consistency loss, which attempts to make the network reproduce similar results independently of the number of steps used to reach the target. We have incorporated this loss into a new landmark-guided face synthesis, coined GANnotation, which allows for high-quality image synthesis even from low resolution images. We showed how the target landmarks become the ground-truth points, thus making GANnotation a powerful tool. We believe this paper opens the research question of pairing distributions even when the results support plausible images. The models used to generate the images of this paper will be made publicly available. 

{\small
\bibliographystyle{ieee}
\bibliography{egbib}

\begin{thebibliography}{10}\itemsep=-1pt

\bibitem{Aly2005}
H.~A. Aly and E.~Dubois.
\newblock Image up-sampling using total-variation regularization with a new
  observation model.
\newblock {\em IEEE Transactions on Image Processing}, 14(10):1647--1659, 2005.

\bibitem{Arjovsky2017}
M.~Arjovsky, S.~Chintala, and L.~Bottou.
\newblock Wasserstein gan.
\newblock {\em arXiv preprint arXiv:1701.07875}, 2017.

\bibitem{Belhumeur13}
P.~Belhumeur, D.~Jacobs, D.~Kriegman, and N.~Kumar.
\newblock Localizing parts of faces using a consensus of exemplars.
\newblock {\em TPAMI}, 35(12):2930--2940, 2013.

\bibitem{Bulat2018}
A.~Bulat and G.~Tzimiropoulos.
\newblock Super-fan: Integrated facial landmark localization and
  super-resolution of real-world low resolution faces in arbitrary poses with
  gans.
\newblock {\em CVPR}, 2018.

\bibitem{Choi2018}
Y.~Choi, M.~Choi, M.~Kim, J.-W. Ha, S.~Kim, and J.~Choo.
\newblock Stargan: Unified generative adversarial networks for multi-domain
  image-to-image translation.
\newblock In {\em CVPR}, 2018.

\bibitem{Di2018}
X.~Di, V.~A. Sindagi, and V.~M. Patel.
\newblock Gp-gan: Gender preserving gan for synthesizing faces from landmarks.
\newblock In {\em ICPR}, 2018.

\bibitem{Goodfellow2014}
I.~Goodfellow, J.~Pouget-Abadie, M.~Mirza, B.~Xu, D.~Warde-Farley, S.~Ozair,
  A.~Courville, and Y.~Bengio.
\newblock Generative adversarial nets.
\newblock In {\em NIPS}, 2014.

\bibitem{Gross10}
R.~Gross, I.~Matthews, J.~Cohn, T.~Kanade, and S.~Baker.
\newblock Multi-pie.
\newblock {\em IMAVIS}, 28(5):807--813, 2010.

\bibitem{Gulrajani2017}
I.~Gulrajani, F.~Ahmed, M.~Arjovsky, V.~Dumoulin, and A.~C. Courville.
\newblock Improved training of wasserstein gans.
\newblock In {\em NIPS}, 2017.

\bibitem{He2016}
K.~He, X.~Zhang, S.~Ren, and J.~Sun.
\newblock Deep residual learning for image recognition.
\newblock In {\em CVPR}, 2016.

\bibitem{Hu2018}
Y.~Hu, X.~Wu, B.~Yu, R.~He, and Z.~Sun.
\newblock Pose-guided photorealistic face rotation.
\newblock In {\em CVPR}, 2018.

\bibitem{Huang2017}
R.~Huang, S.~Zhang, T.~Li, R.~He, et~al.
\newblock Beyond face rotation: Global and local perception gan for
  photorealistic and identity preserving frontal view synthesis.
\newblock In {\em ICCV}, 2017.

\bibitem{Isola2017}
P.~Isola, J.-Y. Zhu, T.~Zhou, and A.~A. Efros.
\newblock Image-to-image translation with conditional adversarial networks.
\newblock {\em CVPR}, 2017.

\bibitem{Johnson2016}
J.~Johnson, A.~Alahi, and L.~Fei-Fei.
\newblock Perceptual losses for real-time style transfer and super resolution.
\newblock In {\em ECCV}, 2016.

\bibitem{Kingma2015}
D.~P. Kingma and J.~Ba.
\newblock Adam: A method for stochastic optimization.
\newblock {\em ICLR}, 2015.

\bibitem{Kossaifi2018}
J.~Kossaifi, L.~Tran, Y.~Panagakis, and M.~Pantic.
\newblock Gagan: Geometry-aware generative adversarial networks.
\newblock In {\em CVPR}, 2018.

\bibitem{Le12}
V.~Le, J.~Brandt, Z.~Lin, L.~D. Bourdev, and T.~S. Huang.
\newblock Interactive facial feature localization.
\newblock In {\em ECCV}, pages 679--692, 2012.

\bibitem{Ledig2017}
C.~Ledig, L.~Theis, F.~Husz{\'a}r, J.~Caballero, A.~Cunningham, A.~Acosta,
  A.~P. Aitken, A.~Tejani, J.~Totz, Z.~Wang, et~al.
\newblock Photo-realistic single image super-resolution using a generative
  adversarial network.
\newblock In {\em CVPR}, 2017.

\bibitem{Lim2017}
J.~H. Lim and J.~C. Ye.
\newblock Geometric gan.
\newblock {\em arXiv preprint arXiv:1705.02894}, 2017.

\bibitem{Liu2015}
Z.~Liu, P.~Luo, X.~Wang, and X.~Tang.
\newblock Deep learning face attributes in the wild.
\newblock In {\em ICCV}, 2015.

\bibitem{Ma2017}
L.~Ma, X.~Jia, Q.~Sun, B.~Schiele, T.~Tuytelaars, and L.~V. Gool.
\newblock Pose guided person image generation.
\newblock In {\em NIPS}, 2017.

\bibitem{Maas2013}
A.~L. Maas, A.~Y. Hannun, and A.~Y. Ng.
\newblock Rectifier nonlinearities improve neural network acoustic models.
\newblock In {\em ICML Workshop on Deep Learning for Audio, Speech and Language
  Processing}, 2013.

\bibitem{Mao2017}
X.~Mao, Q.~Li, H.~Xie, R.~Y. Lau, Z.~Wang, and S.~P. Smolley.
\newblock Least squares generative adversarial networks.
\newblock In {\em ICCV}, 2017.

\bibitem{Miyato2018}
T.~Miyato, T.~Kataoka, M.~Koyama, and Y.~Yoshida.
\newblock Spectral normalization for generative adversarial networks.
\newblock In {\em ICLR}, 2018.

\bibitem{Paszke2017}
A.~Paszke, S.~Gross, S.~Chintala, G.~Chanan, E.~Yang, Z.~DeVito, Z.~Lin,
  A.~Desmaison, L.~Antiga, and A.~Lerer.
\newblock Automatic differentiation in pytorch.
\newblock 2017.

\bibitem{Pumarola2018}
A.~Pumarola, A.~Agudo, A.~Martinez, A.~Sanfeliu, and F.~Moreno-Noguer.
\newblock Ganimation: Anatomically-aware facial animation from a single image.
\newblock In {\em ECCV}, 2018.

\bibitem{Qiao2018}
F.~Qiao, N.-M. Yao, Z.~Jiao, Z.~Li, H.~Chen, and H.~Wang.
\newblock Geometry-contrastive generative adversarial network for facial
  expression synthesis.
\newblock {\em CoRR}, abs/1802.01822, 2018.

\bibitem{Radford2015}
A.~Radford, L.~Metz, and S.~Chintala.
\newblock Unsupervised representation learning with deep convolutional
  generative adversarial networks.
\newblock In {\em ICLR}, 2016.

\bibitem{Sagonas13}
C.~Sagonas, G.~Tzimiropoulos, S.~Zafeiriou, and M.~Pantic.
\newblock A semi-automatic methodology for facial landmark annotation.
\newblock In {\em CVPR'W}, 2013.

\bibitem{Shen15}
J.~Shen, S.~Zafeiriou, G.~S. Chrysos, J.~Kossaifi, G.~Tzimiropoulos, and
  M.~Pantic.
\newblock The first facial landmark tracking in-the-wild challenge: Benchmark
  and results.
\newblock In {\em ICCV'W}, 2015.

\bibitem{Simonyan15}
K.~Simonyan and A.~Zisserman.
\newblock Very deep convolutional networks for large-scale image recognition.
\newblock In {\em ICLR}, 2015.

\bibitem{Tian2018}
Y.~Tian, X.~Peng, L.~Zhao, S.~Zhang, and D.~N. Metaxas.
\newblock Cr-gan: Learning complete representations for multi-view generation.
\newblock {\em IJCAI}, 2018.

\bibitem{Valstar2015}
M.~F. Valstar, T.~Almaev, J.~M. Girard, G.~McKeown, M.~Mehu, L.~Yin, M.~Pantic,
  and J.~F. Cohn.
\newblock Fera 2015 - second facial expression recognition and analysis
  challenge.
\newblock 2015.

\bibitem{Wang2018}
W.~Wang, X.~Alameda-Pineda, D.~Xu, P.~Fua, E.~Ricci, and N.~Sebe.
\newblock Every smile is unique: Landmark-guided diverse smile generation.
\newblock In {\em CVPR}, 2018.

\bibitem{Wang2018non}
X.~Wang, R.~Girshick, A.~Gupta, and K.~He.
\newblock Non-local neural networks.
\newblock In {\em CVPR}, 2018.

\bibitem{Wu2018}
X.~Wu, R.~He, Z.~Sun, and T.~Tan.
\newblock A light cnn for deep face representation with noisy labels.
\newblock {\em IEEE Transactions on Information Forensics and Security},
  13(11):2884--2896, 2018.

\bibitem{Yin2017}
X.~Yin, X.~Yu, K.~Sohn, X.~Liu, and M.~Chandraker.
\newblock Towards large-pose face frontalization in the wild.
\newblock In {\em ICCV}, 2017.

\bibitem{Zhang2018}
H.~Zhang, I.~Goodfellow, D.~Metaxas, and A.~Odena.
\newblock Self-attention generative adversarial networks.
\newblock {\em arXiv preprint arXiv:1805.08318}, 2018.

\bibitem{Zhang2014bp4d}
X.~Zhang, L.~Yin, J.~F. Cohn, S.~Canavan, M.~Reale, A.~Horowitz, P.~Liu, and
  J.~M. Girard.
\newblock Bp4d-spontaneous: a high-resolution spontaneous 3d dynamic facial
  expression database.
\newblock {\em Image and Vision Computing}, 32(10):692--706, 2014.

\bibitem{Zhao2017}
J.~Zhao, L.~Xiong, P.~Karlekar~Jayashree, J.~Li, F.~Zhao, Z.~Wang,
  P.~Sugiri~Pranata, P.~Shengmei~Shen, S.~Yan, and J.~Feng.
\newblock Dual-agent gans for photorealistic and identity preserving profile
  face synthesis.
\newblock In {\em NIPS}, 2017.

\bibitem{Zhu2017}
J.-Y. Zhu, T.~Park, P.~Isola, and A.~A. Efros.
\newblock Unpaired image-to-image translation using cycle-consistent
  adversarial networks.
\newblock In {\em ICCV}, 2017.

\bibitem{Zhu12}
X.~Zhu and D.~Ramanan.
\newblock Face detection, pose estimation, and landmark localization in the
  wild.
\newblock In {\em CVPR}, pages 2879--2886, 2012.

\end{thebibliography}
}
\end{document}